\documentclass[letterpaper, 10 pt, conference]{IEEEtran}
\usepackage{times}

\usepackage[numbers]{natbib}
\usepackage{multicol}
\usepackage[bookmarks=true]{hyperref}

\usepackage{amssymb}
\usepackage{amsmath}

\usepackage{graphicx}
\usepackage{subcaption}
\usepackage{booktabs}

\usepackage{array}

\begin{document}

\title{\huge \vspace{0.5cm} Real-Time Trajectory Replanning for MAVs using Uniform B-splines and a 3D Circular Buffer}

\author{Vladyslav Usenko, Lukas von Stumberg, Andrej Pangercic and Daniel Cremers \\ Technical University of Munich }



%

\maketitle

\begin{abstract}
In this paper, we present a real-time approach to local trajectory replanning for microaerial vehicles (MAVs). Current  trajectory generation methods for multicopters achieve high success rates in cluttered environments, but assume that the environment is static and require prior knowledge of the map. In the presented study, we use the results of such planners and extend them with a local replanning algorithm that can handle unmodeled (possibly dynamic) obstacles while keeping the MAV close to the global trajectory. To ensure that the proposed approach is real-time capable, we maintain information about the environment around the MAV in an occupancy grid stored in a three-dimensional circular buffer, which moves together with a drone, and represent the trajectories by using uniform B-splines. This representation ensures that the trajectory is sufficiently smooth and simultaneously allows for efficient optimization.
\end{abstract}

\IEEEpeerreviewmaketitle

\section{Introduction}
\label{sec:introduction}


In recent years, microaerial vehicles (MAVs) have gained popularity in many practical applications such as aerial photography, inspection, surveillance and even delivery of goods. Most commercially available drones assume that the path planned by the user is collision-free or provide only limited obstacle-avoidance capabilities. To ensure safe navigation in the presence of unpredicted obstacles a replanning method that generates a collision-free trajectory is required.

Formulation of the trajectory generation problem largely depends on the application and assumptions about the environment. In the case where an MAV is required to navigate a cluttered environment, possibly an indoor one, we would suggest subdividing the problem into two layers. First, we assume that a map of the environment is available and a trajectory from a specified start point to the goal point is planned in advance. 

This task has been a popular research topic in recent years, and several solutions have been proposed by \citet{achtelik2014motion} and \citet{richter2016polynomial}. They used occupancy representation of the environment to check for collisions and searched for the valid path in a visibility graph constructed using sampling based planners. Thereafter, they followed the approach proposed by \citet{mellinger2011trajectory} to fit polynomial splines through the points of the planned path to generate a smooth feasible trajectory. The best algorithms of this type can compute a trajectory through tens of waypoints in several seconds.

\begin{figure}
		\centering
        \begin{subfigure}[b]{0.8\linewidth}
                \includegraphics[width=\textwidth]{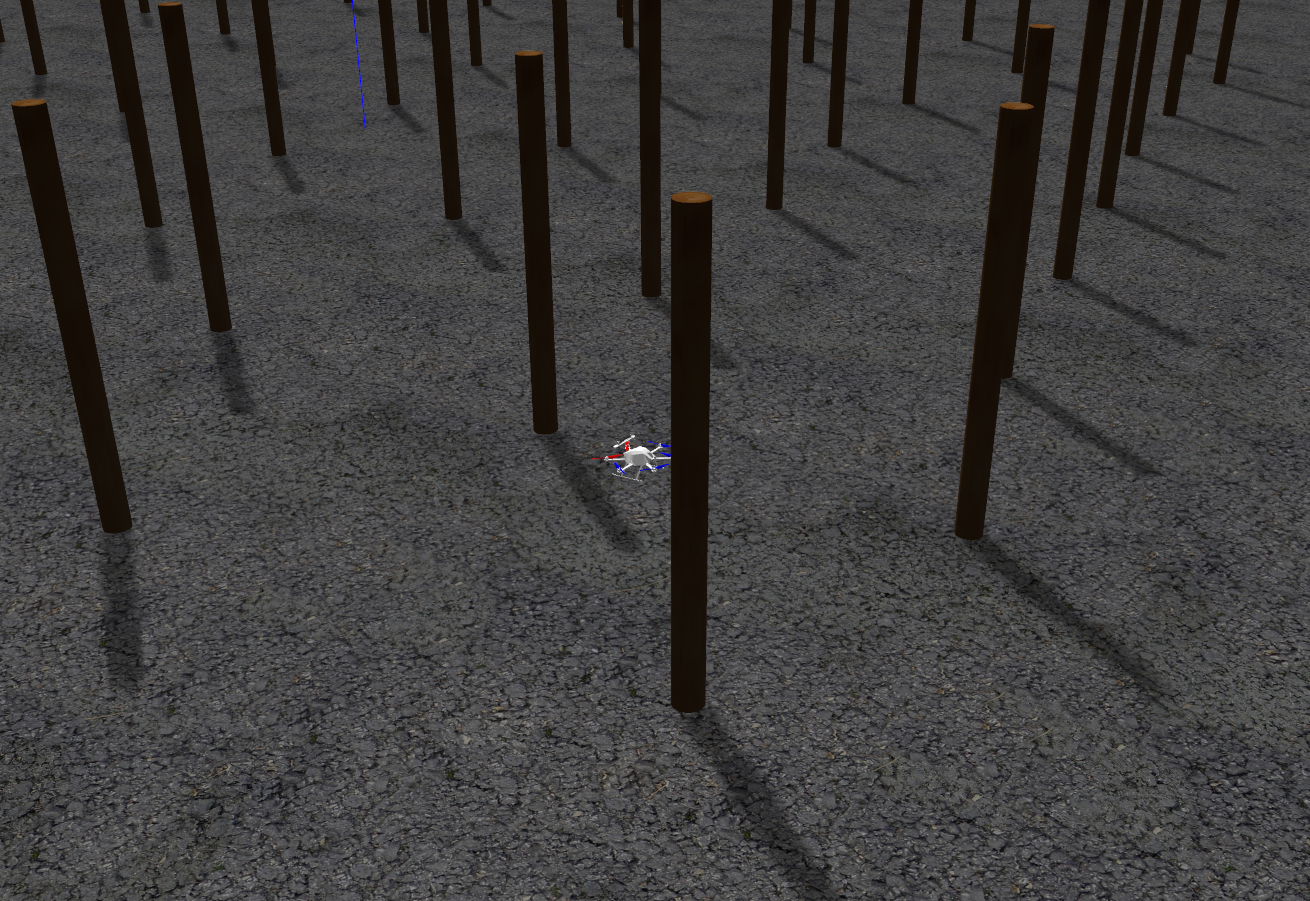}
        \end{subfigure}
        \begin{subfigure}[b]{0.8\linewidth}
                \includegraphics[width=\textwidth]{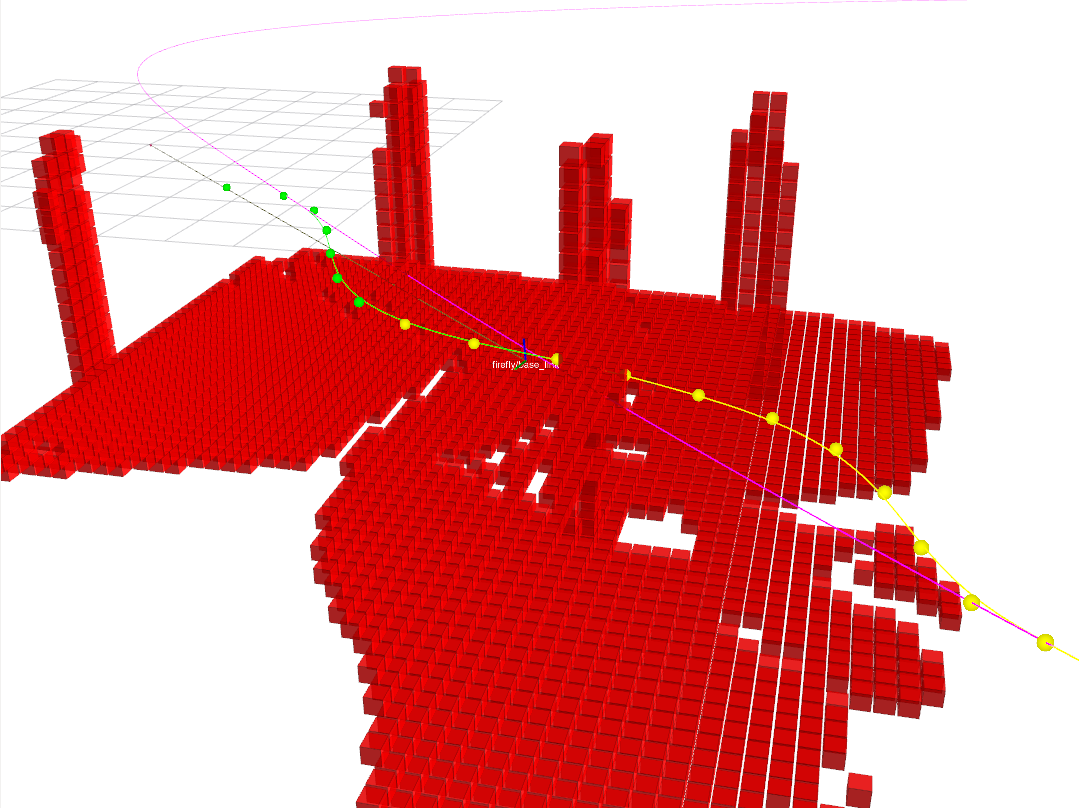}
        \end{subfigure}
        \caption{Example of local trajectory replanning algorithm running in the simulator. Global trajectory is visualized in purple and the local obstacle map is visualized in red. The local trajectory is represented by a uniform quintic B-spline, and its control points are visualized in yellow for the fixed parts and in green for the parts that can still change due to optimization.}
        \label{fig:simulation}
\end{figure}

To cope with any unmodeled, possibly dynamic, obstacle a lower planning level is required, which can generate a trajectory that keeps the MAV close to the global path and simultaneously avoids unpredicted obstacles based on an environment representation constructed from the most recent sensor measurements. This replanning level should run in several milliseconds to ensure the safety of MAVs operating at high velocities.

The proposed approach solves a similar problem as that solved by \citet{oleynikova2016continuous-time}, but instead of using polynomial splines for representing the trajectory we propose the use of B-splines and discuss their advantages over polynomial splines for this task. Furthermore, we propose the use of a robocentric, fixed-size three-dimensional (3D) circular buffer to maintain local information about the environment. Even though the proposed method cannot model arbitrarily large occupancy maps, as some octree implementations, faster look-up and measurement insertion operations make it better suited for real-time replanning tasks. 

We demonstrate the performance of the system in several simulated and real-world experiments, and provide open-source implementation of the software to community.

The contributions of the present study are as follows:
\begin{itemize}
\item Formulation of local trajectory replanning as a B-spline optimization problem and thorough comparison with alternative representations (polynomial, discrete).
\item High-performance 3D circular buffer implementation for local obstacle mapping and collision checking and comparison with alternative methods.
\item System design and evaluation on realistic simulator and real hardware, in addition to performance comparison with existing methods.
\end{itemize}

In addition to analyzing the results presented in the paper, we encourage the reader to watch the demonstration video and inspect the available code, which can be found at

\begin{center}
\urlstyle{tt}
\url{https://vision.in.tum.de/research/robotvision/replanning}
\end{center}

\section{Related Work}
\label{sec:related_work}

%
%

In this section, we describe the studies relevant to different aspects of collision-free trajectory generation. First, we discuss existing trajectory generation strategies and their applications to MAV motion planning. Thereafter, we discuss the state-of-the-art approaches for 3D environment mapping.

\subsection{Trajectory generation}
Trajectory generation strategies can be subdivided into three main approaches: search-based path planning followed by smoothing, optimization-based approaches and motion-primitive-based approaches.

In search-based approaches, first, a non-smooth path is constructed on a graph that represents the environment. This graph can be a fully connected grid as in \cite{dolgov2008practical} and \cite{jung2008line}, or be computed using a sampling-based planner (RRT, PRM) as in \cite{richter2016polynomial} and \cite{burri2015real-time}. Thereafter, a smooth trajectory represented by a polynomial, B-spline or discrete set of points is computed to closely follow this path. This class of approaches is currently the most popular choice for large-scale path planning problems in cluttered environments where a map is available a priory.

Optimization-based approaches rely on minimizing a cost function that consists of smoothness and collision terms. The trajectory itself can be represented as a set of discrete points \cite{zucker2013chomp} or polynomial segments \cite{oleynikova2016continuous-time}. The approach presented in the present work falls into this category, but represents a trajectory using uniform B-splines. 

Another group of approaches is based on path sampling and motion primitives. Sampling-based approaches were successfully used for challenging tasks such as ball juggling \cite{mueller2013computationally}, and motion primitives were successfully applied to flight through the forest \cite{paranjape2015motion}, but the ability of both approaches to find a feasible trajectory  depends largely on the selected discretization scheme. 

\subsection{Environment representation}

To be able to plan a collision-free trajectory a representation of the environment that stores information about occupancy is required. The simplest solution that can be used in the 3D case is a voxel grid. In this representation, a volume is subdivided into regular grid of smaller sub-volumes (voxels), where each voxel stores information about its occupancy. The main drawback of this approach is its large memory-footprint, which allows for maping only small fixed-size volumes. The advantage, however, is very fast constant time access to any element.

To deal with the memory limitation, octree-based representations of the environment are used in \cite{hornung13auro} \cite{steinbrucker2014volumetric}. They store information in an efficient way by pruning the leaves of the trees that contain the same information, but the access times for each element become logarithmic in the number of nodes, instead of the constant time as in the voxel-based approaches.

Another popular approach to environment mapping is voxel hashing, which was proposed by \citet{niessner2013real} and used in \cite{oleynikova2016voxblox}. It is mainly used for storing a truncated signed distance function representation of the environment. In this case, only a narrow band of measurements around the surface is inserted and only the memory required for that sub-volume is allocated. However, when full measurements must be inserted or the dense information must be stored the advantages of this approach compared to those of the other approaches are not significant.

\begin{figure*}
		\begin{subfigure}[b]{0.245\linewidth}
                \includegraphics[width=\textwidth]{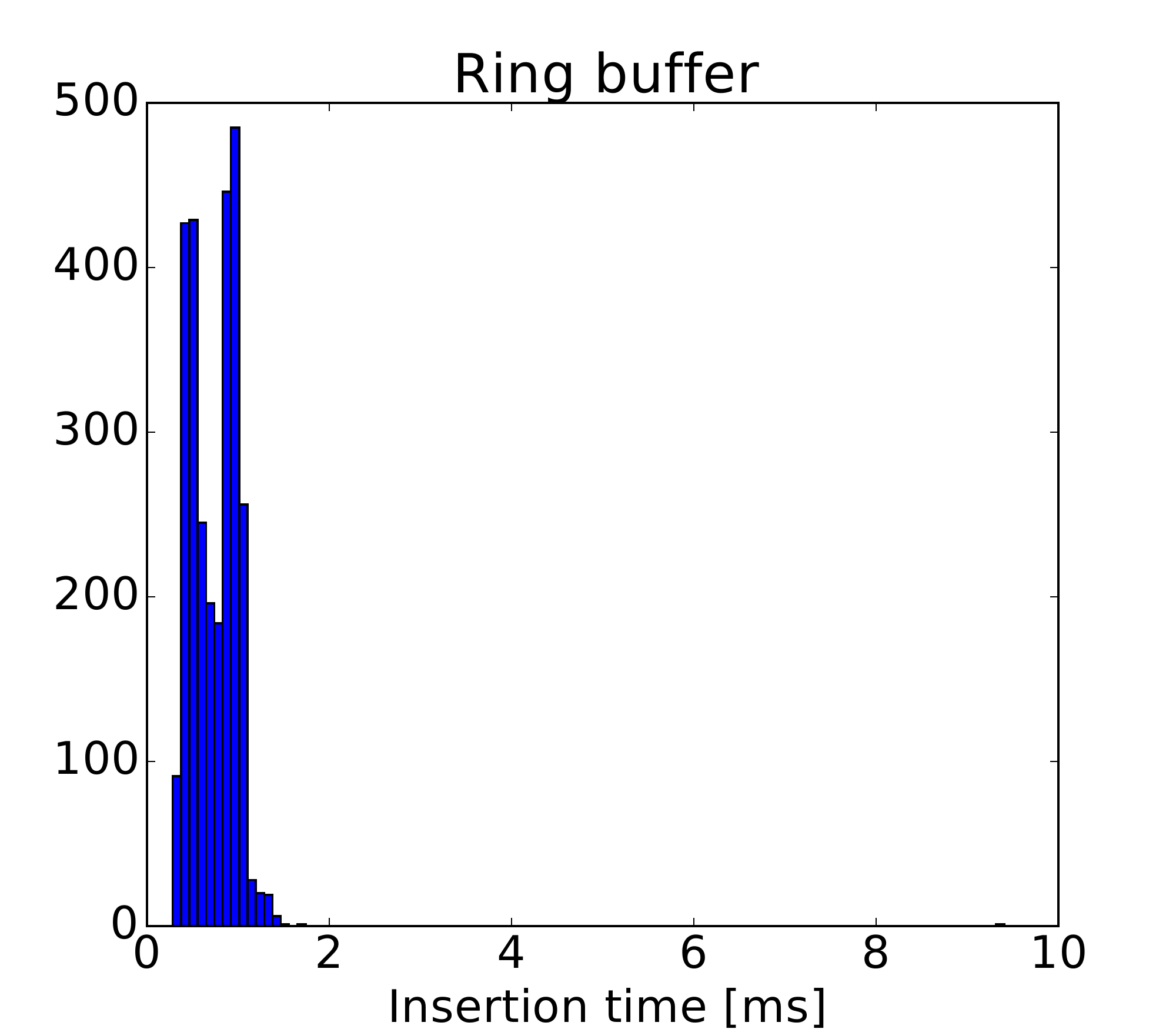}
                \caption{}
                \label{fig:ring_buffer_hist}
        \end{subfigure}
        \begin{subfigure}[b]{0.245\linewidth}
                \includegraphics[width=\textwidth]{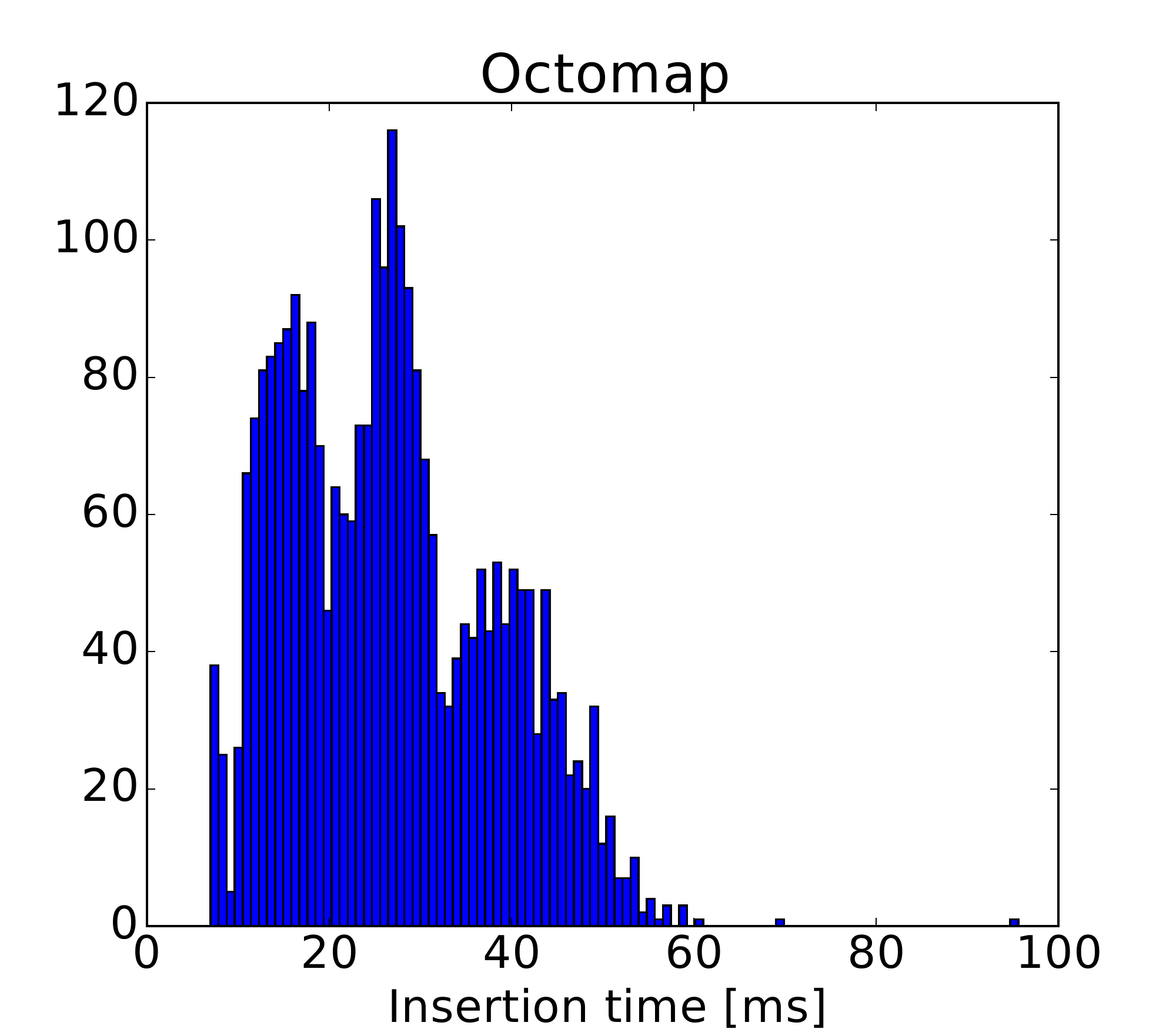}
                \caption{}
                \label{fig:octomap_hist}
        \end{subfigure}
        \begin{subfigure}[b]{0.245\linewidth}
                \includegraphics[width=\textwidth]{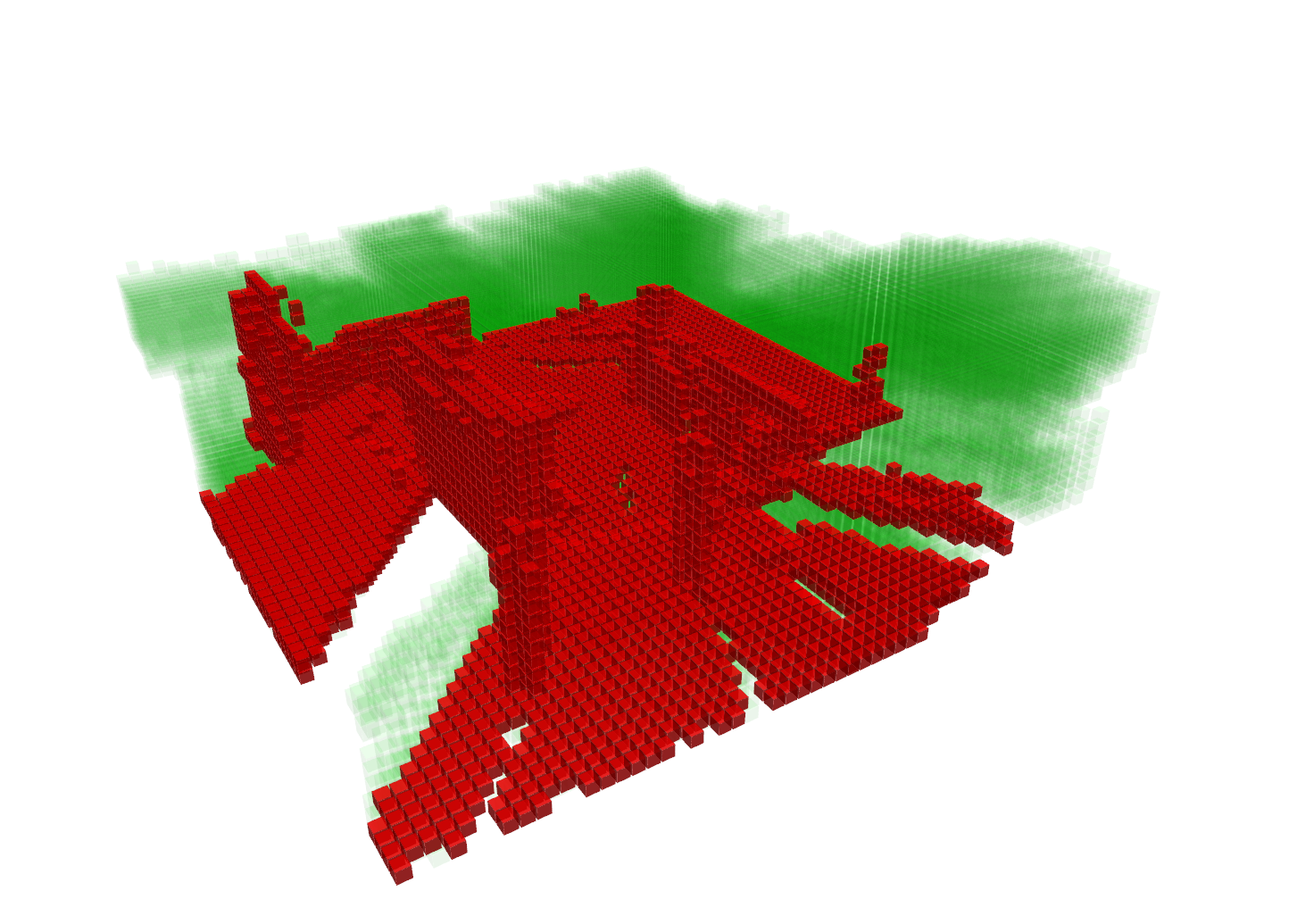}
                \caption{}
                \label{fig:ring_buffer}
        \end{subfigure}
        \begin{subfigure}[b]{0.245\linewidth}
                \includegraphics[width=\textwidth]{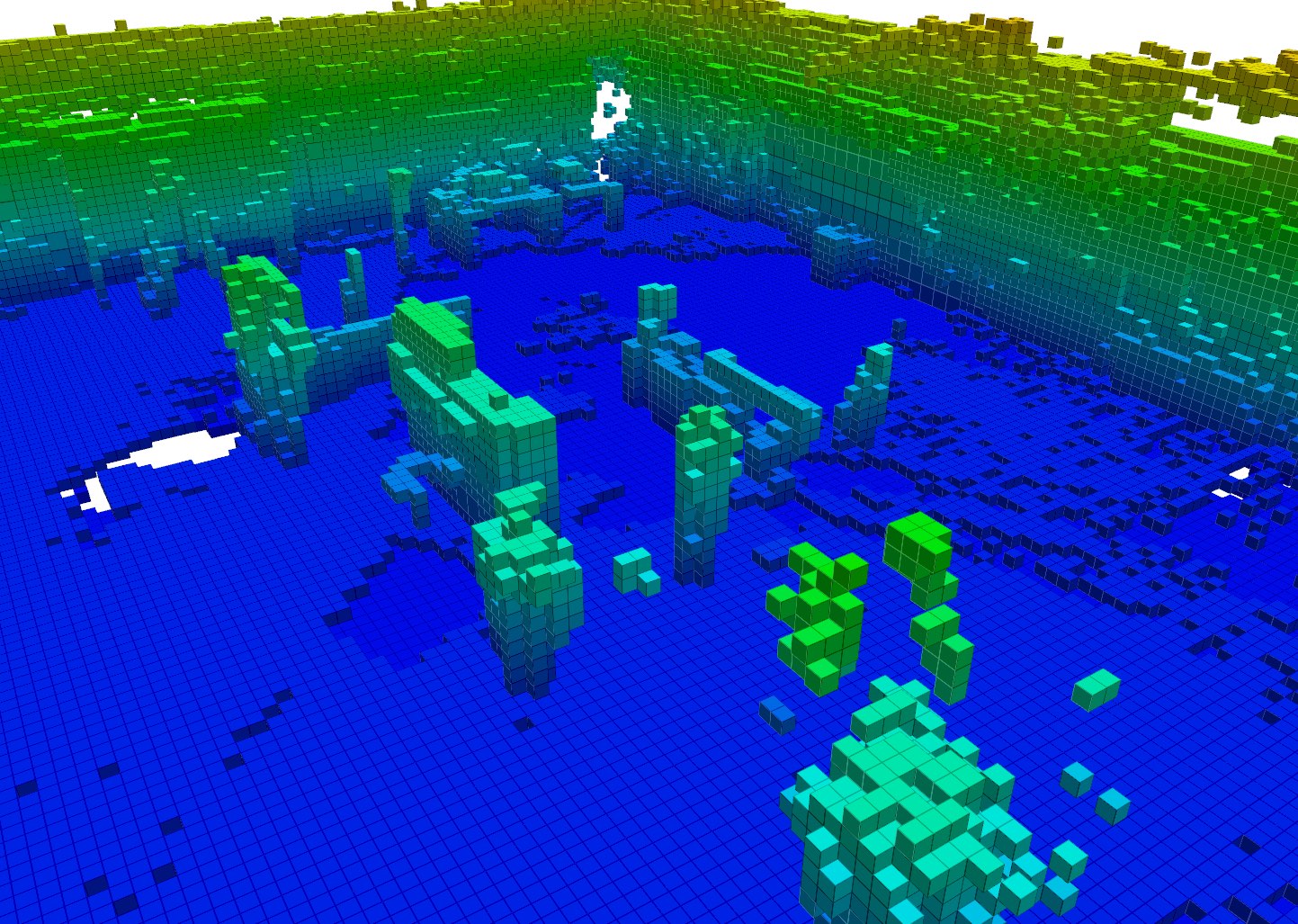}
                \caption{}
                \label{fig:octomap}
        \end{subfigure}
        \caption{Comparison between octomap and circular buffer for occupancy mapping on fr2/pioneer\_slam2 sequence of \cite{sturm12iros}. Being able to map only a local environment around the robot (3 m at voxel resolution of 0.1 m) circular buffer is more than an order of magnitude faster when inserting point cloud measurements from a depth map subsampled to a resolution of  $160 \times 120$. Subplots (\subref{fig:ring_buffer_hist}) and (\subref{fig:octomap_hist}) show the histograms of insertion time, and (\subref{fig:ring_buffer}) and (\subref{fig:octomap}) show the qualitative results of the circular buffer (red: occupied, green:free) and the octomap, respectively.}
        \label{fig:octomap_ring_buffer}
\end{figure*}

\section{Trajectory Representation using uniform B-splines}
\label{sec:b_splines}

We use uniform B-spline representation for the trajectory function $p(t)$. Because, as shown in \cite{mellinger2011trajectory} and \cite{achtelik2014motion}, the trajectory must be continuous up to the forth derivative of position (snap), we use quintic B-splines to ensure the required smoothness of the trajectory.

\subsection{Uniform B-splines}

The value of a B-spline of degree $k-1$ can be evaluated using the following equation:
\begin{align}
p(t) = \sum_{i=0}^{n} p_i B_{i,k}(t),
\end{align}
where $p_i \in \mathbb{R}^n$ are control points at times $t_i, i \in [0, .., n]$ and $B_{i,k}(t)$ are basis functions that can be computed using the De Boor -- Cox recursive formula \cite{de1972calculating} \cite{cox1972numerical}. Uniform B-splines have a fixed time interval $\Delta t$ between their control points, which simplifies computation of the basis functions.

In the case of quintic uniform B-splines, at time $t \in [t_i, t_{i+1})$ the value of $p(t)$ depends only on six control points, namely $[t_{i-2}, t_{i-1}, t_i, t_{i+1}, t_{i+2}, t_{i+3}]$. To simplify calculations we transform time to a uniform representation $s(t) = (t - t_0)/\Delta t$, such that the control points transform into $s_i \in [0,..,n]$. We define function $u(t) = s(t)-s_i$ as time elapsed since the start of the segment. Following the matrix representation of the De Boor -- Cox formula \cite{qin}, the value of the function can be evaluated as follows:

\begin{align}
    p(u(t)) &= 
    \begin{pmatrix}
    1 \\
    u \\
    u^2 \\
    u^3 \\
    u^4 \\
    u^5 \\
    \end{pmatrix}^T
    M_6
    \begin{pmatrix}
    p_{i-2}\\
    p_{i-1}\\
    p_{i}\\
    p_{i+1}\\
    p_{i+2}\\
    p_{i+3}\\
    \end{pmatrix},
\end{align}

\begin{align}
    M_6 = \frac{1}{5!}
    \begin{pmatrix}
    1 &  26 &  66 & 26 &  1 &  0 \\
  -5 & -50 &   0 &  50 &   5 &   0 \\
  10 &  20 & -60 &  20 &  10 &   0 \\
-10 &  20 &  0 & -20 &  10 &  0 \\
   5 & -20 &  30 & -20 &  5 &   0 \\
  -1 &  5 & -10 & 10 &  -5 & 1 
    \end{pmatrix}.
\end{align}

Given this formula, we can compute derivatives with respect to time (velocity, acceleration) as follows:
\begin{align}
    p'(u(t)) &= \frac{1}{\Delta t}
    \begin{pmatrix}
    0 \\
    1 \\
    2u \\
    3u^2 \\
    4u^3 \\
    5u^4 \\
    \end{pmatrix}^T
    M_6
    \begin{pmatrix}
    p_{i-2}\\
    p_{i-1}\\
    p_{i}\\
    p_{i+1}\\
    p_{i+2}\\
    p_{i+3}\\
    \end{pmatrix},
\end{align}

\begin{align}
    p''(u(t)) &= \frac{1}{\Delta t^2}
    \begin{pmatrix}
    0 \\
    0 \\
    2 \\
    6u \\ 
    12u^2 \\
    20u^3 \\
    \end{pmatrix}^T
    M_6
    \begin{pmatrix}
    p_{i-2}\\
    p_{i-1}\\
    p_{i}\\
    p_{i+1}\\
    p_{i+2}\\
    p_{i+3}\\
    \end{pmatrix}.
\end{align}

The computation of other time derivatives and derivatives with respect to control points is also straightforward.

The integral over squared time derivatives can be computed in the closed form. For example, the integral over squared acceleration can be computed as follows:
\begin{align}
    E_q &=
    \int_{t_i}^{t_{i+1}} p''(u(t))^2 dt \\
    &=
    \begin{pmatrix}
    p_{i-2}\\
    p_{i-1}\\
    p_{i}\\
    p_{i+1}\\
    p_{i+2}\\
    p_{i+3}\\
    \end{pmatrix}^T
    M_6^T
    Q
    M_6
    \begin{pmatrix}
    p_{i-2}\\
    p_{i-1}\\
    p_{i}\\
    p_{i+1}\\
    p_{i+2}\\
    p_{i+3}\\
    \end{pmatrix}, \\
\end{align}
where
\begin{align}
    Q &= \frac{1}{\Delta t^3} \int_0^1
    \begin{pmatrix}
    0 \\
    0 \\
    2 \\
    6u \\ 
    12u^2 \\
    20u^3 \\
    \end{pmatrix}
    \begin{pmatrix}
    0 \\
    0 \\
    2 \\
    6u \\ 
    12u^2 \\
    20u^3 \\
    \end{pmatrix}^T du \\
    &= \frac{1}{\Delta t^3} 
    \begin{pmatrix}
      0   &    0    &   0    &   0   &    0    &   0 \\
      0   &    0    &  0     &  0   &    0  &     0 \\
      0   &    0   &    8   &   12  &    16  &    20 \\
      0   &    0   &   12   &   24  &    36   &   48 \\
      0   &    0   &   16   &  36  &  57.6   &   80 \\
      0   &    0   &   20   &   48   &   80 & 114.286 
    \end{pmatrix}.
\end{align}
Please note that matrix Q is constant in the case of uniform B-splines. Therefore, it can be computed in advance for determining the integral over any squared derivative (see \cite{richter2016polynomial} for details).

\subsection{Comparison with polynomial trajectory representation}
In this subsection, we discuss the advantages and disadvantages of B-spline trajectory representation compared to polynomial-splines-based representation \cite{richter2016polynomial} \cite{oleynikova2016continuous-time}.


To obtain a trajectory that is continuous up to the forth derivative of position, we need to use B-splines of degree five or greater and polynomial splines of at least degree nine (we need to set five boundary constraints on each endpoint of the segment). Furthermore, for polynomial splines we must explicitly include boundary constraints into optimization, while the use of B-splines guarantees the generation of a smooth trajectory for an arbitrary set of control points. Another useful property of B-splines is the locality of trajectory changes due to changes in the control points, which means that a change in one control point affects only a few segments in the entire trajectory. All these properties result in faster optimization because there are fewer variables to optimize and fewer constraints.

Evaluation of position at a given time, derivatives with respect to time (velocity, acceleration, jerk, snap), and integrals over squared time derivatives are similar for both cases because closed-form solutions are available for both cases.

The drawback of B-splines, however, is the fact that the trajectory does not pass through the control points. This makes it difficult to enforce boundary constraints. The only constraint we can enforce is a static one (all time derivatives are zero), which can be achieved by inserting the same control point $k+1$ times, where $k$ is the degree of the B-spline. If we need to set an endpoint of the trajectory to have a non-zero time derivative, an iterative optimization algorithm must be used.

These properties make polynomial splines more suitable for the cases where the control points come from planning algorithms (RRT, PRM), which means that the trajectory must pass through them, else the path cannot be guaranteed to be collision-free. For local replanning, which must account for unmodeled obstacles, this property is not very important; thus, the use of B-spline trajectory representation is a better option.


\begin{figure}
	    \centering
	    \includegraphics[width=0.75\linewidth]{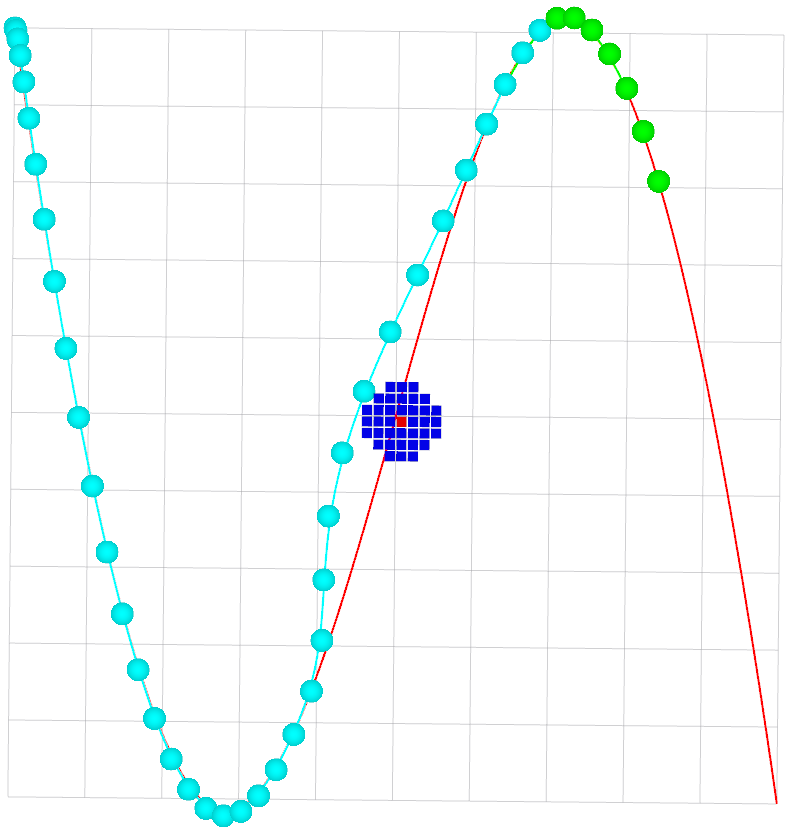}
        \caption{Example of online trajectory replanning using proposed optimization objective. The plot shows a global trajectory computed by fitting a polynomial spline through fixed waypoints (red), voxels within 0.5 m of the obstacle (blue), computed B-spline trajectory with fixed (cyan) and still optimized (green) segments and control points. In the areas with no obstacles, the computed trajectory closely follows the global one, while close to an obstacle, the proposed method generates a smooth trajectory that avoids the obstacle, and rejoins the global trajectory.}
        \label{fig:optimization_example}
\end{figure}

\begin{figure*}
		\begin{subfigure}[b]{0.33\linewidth}
                \includegraphics[width=\textwidth]{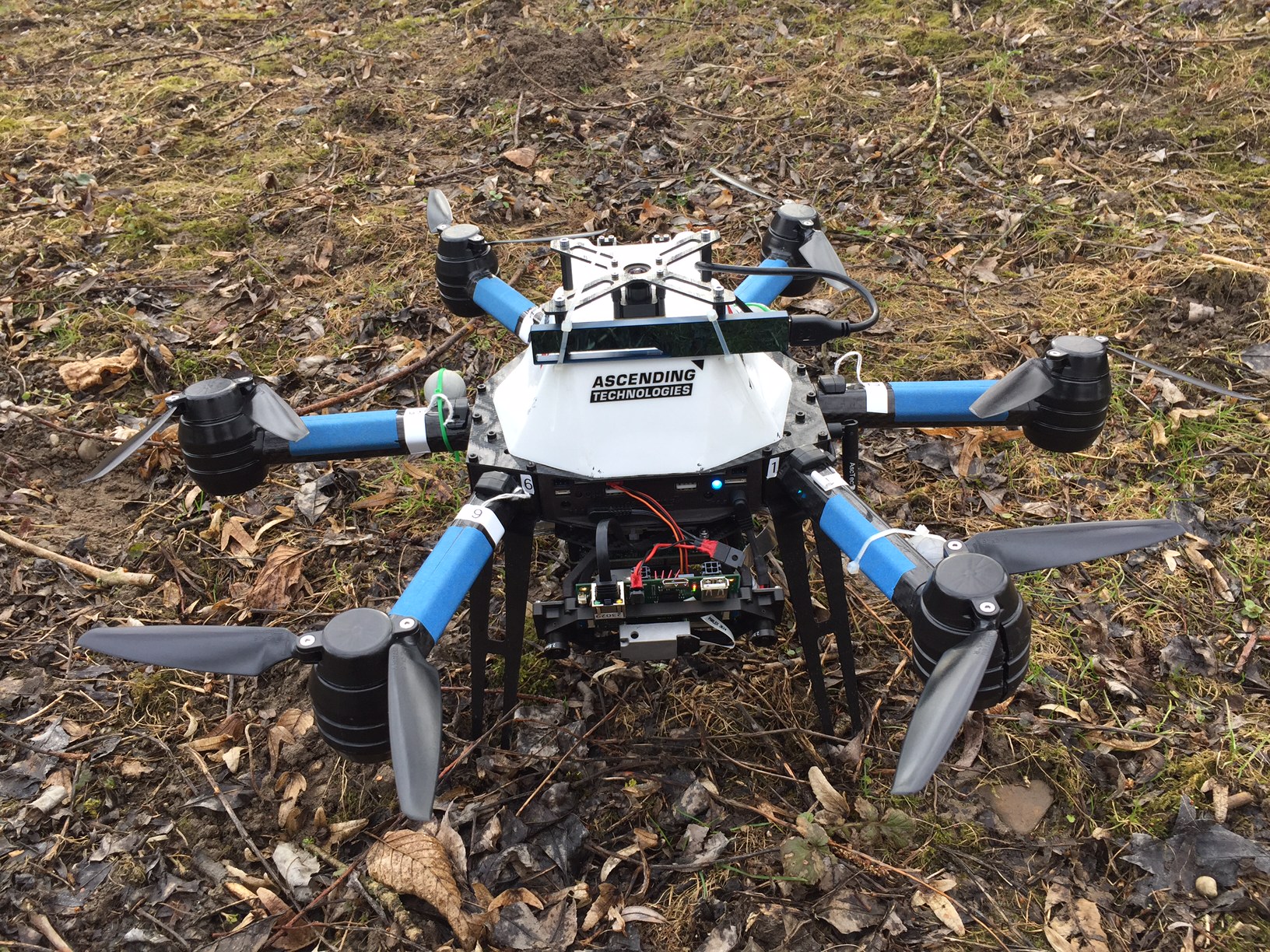}
                \caption{}
                \label{fig:live_test_drone}
        \end{subfigure}
        \begin{subfigure}[b]{0.33\linewidth}
                \includegraphics[width=\textwidth]{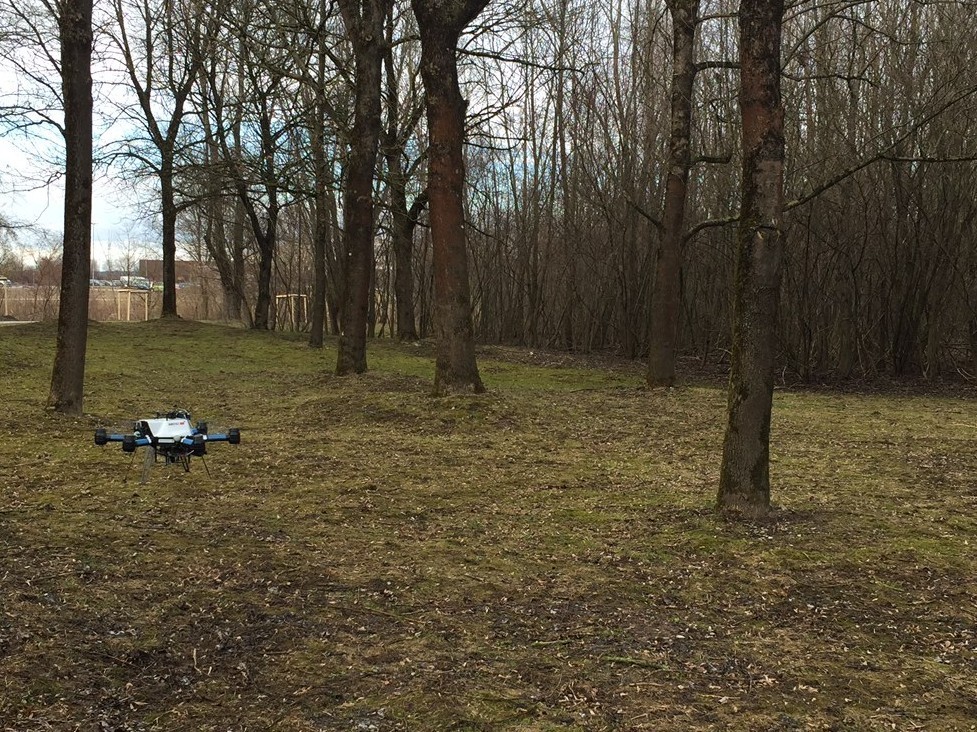}
                \caption{}
                \label{fig:live_test_side_view}
        \end{subfigure}
        \begin{subfigure}[b]{0.33\linewidth}
                \includegraphics[width=\textwidth]{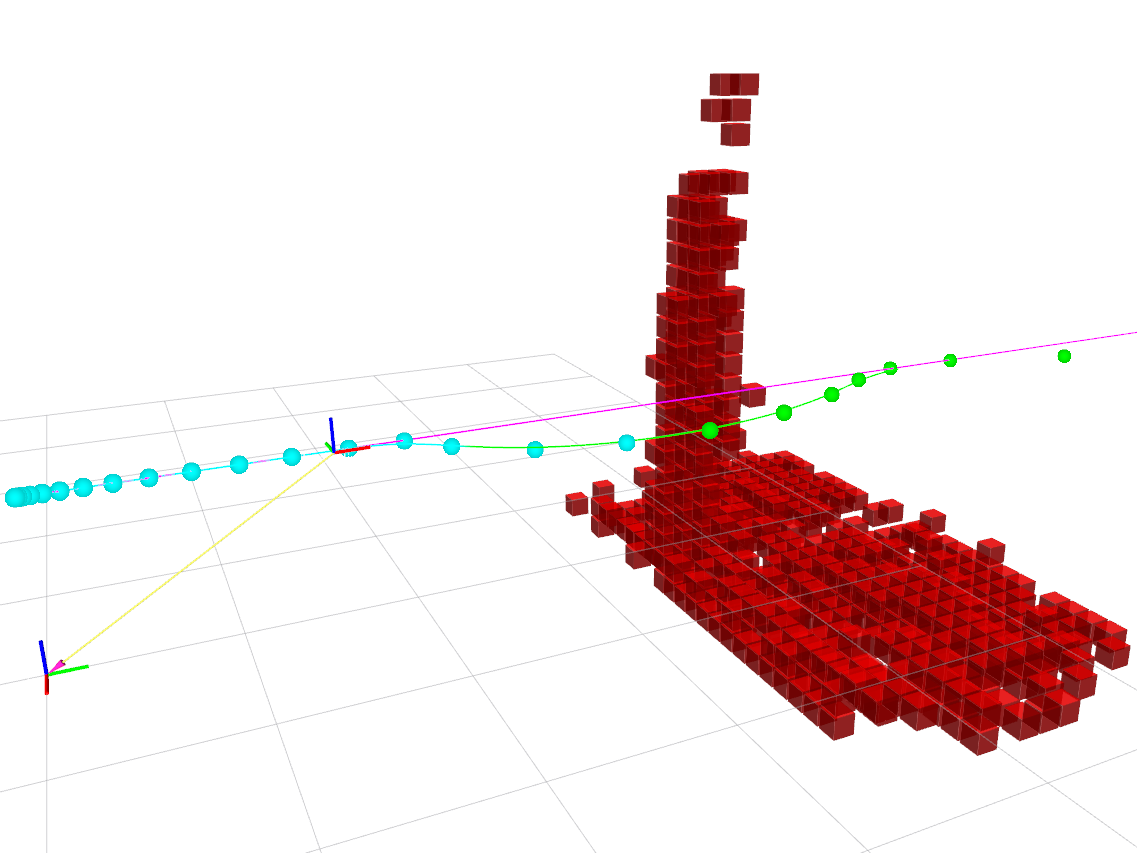}
                \caption{}
                \label{fig:live_test_visualization}
        \end{subfigure}

        \caption{Real-world experiment performed outdoors. The drone (AscTec Neo) equipped with RGB-D camera (Intel Realsense R200) is shown in (\subref{fig:live_test_drone}). In the experiment, the global path is set to a straight line with the goal position 30 m ahead of the drone, and trees act as unmapped obstacles that the drone must avoid. Side view of the scene is shown in (\subref{fig:live_test_side_view}), and visualization with the planned trajectory is shown in (\subref{fig:live_test_visualization}). }
        \label{fig:real_world_experiment}
\end{figure*}

\section{Local Environment Map using 3D Circular Buffer}
\label{sec:circular_buffer}

To to avoid obstacles during flight, we need to maintain an occupancy model of the environment. On one hand, the model should rely on the most recent sensor measurements, and on the other hand it should store some information over time because the field of view of the sensors mounted on the MAV is usually limited.



We argue that for local trajectory replanning a simple solution with a robocentric 3D circular buffer is beneficial. In what follows, we discuss details pertaining to implementation and advantages from the application viewpoint.

\subsection{Addressing}

To enable addressing we discretize the volume into voxels of size $r$. This gives us a mapping from point $p$ in 3D space to an integer valued index $x$ that identifies a particular voxel, and the inverse operation that given an index outputs its center point.


A circular buffer consists of a continuous array of size $N$ and an offset index $o$ that defines the location of the coordinate system of the volume. With this information, we can define the functions to check whether a voxel is in the volume and find its address in the stored array:
\begin{align}
insideVolume(x) &= 0 \le x-o < N, \\
address(x) &= (x-o) \mod N.
\end{align}
If we restrict the size of the array to $N = 2^p$, we can rewrite these functions to use cheap bitwise operations instead of divisions:

\begin{align}
insideVolume(x) &= ~ ! ~( (x-o) ~ \& ~ (\sim(2^p-1))), \\
address(x) &= (x-o) ~ \& ~ (2^p-1).
\end{align}
where $\&$ is a "bitwise and," $\sim$ is a "bitwise negation," and $!$ is a "boolean not.".

To ensure that the volume is centered around the camera, we must simply change the offset $o$ and clear the updated part of the volume. This eliminates the need to copy large amounts of data when the robot moves.

\subsection{Measurement insertion}


We assume that the measurements are performed using range sensors, such as Lidar, RGB-D cameras, and stereo cameras, and can be inserted into the occupancy buffer by using raycast operations.

We use an additional flag buffer to store a set of voxels affected by insertion. First, we iterate over all points in our measurements, and for the points that lie inside the volume, we mark the corresponding voxels as occupied. For the points that lie outside the volume, we compute the closest point inside the volume and mark the corresponding voxels as free rays. Second, we iterate over all marked voxels and perform raycasting toward the sensor origin. We use a 3D variant of \emph{Bresenham's line algorithm} \cite{Amanatides87afast} to increase the efficiency of the raycasting operation.

Thereafter, we again iterate over the volume and update the volume elements by using the hit and miss probabilities, similarly to the approach described in \cite{hornung13auro}.

\subsection{Distance map computation}

To facilitate fast collision checking for the trajectory, we compute the Euclidean distance transform (EDT) of the occupancy volume. This way, a drone approximated by the bounding sphere can be checked for collision by one look-up query. We use an efficient $O(n)$ algorithm written by \citet {felzenszwalb2012distance} to compute EDT of the volume, where $n = N^3$ is the number of voxels in the grid (the complexity is cubic in the size of the volume along a single axis). For querying distance and computing gradient, trilinear interpolation is used.

\section{Trajectory Optimization}
\label{sec:optimization}

The local replanning problem is represented as an optimization of the following cost function:
\begin{align}
E_{total} &=  E_{ep} +  E_{c} +  E_{q} + E_{l},
\end{align}
where $E_{ep}$ is an endpoint cost function that penalizes position and velocity deviations at the end of the optimized trajectory segment from the desired values that usually come from the global trajectory; $E_{c}$ is a collision cost function; $E_{q}$ is the cost of the integral over the squared derivatives (acceleration, jerk, snap); and $E_{l}$ is a soft limit on the norm of time derivatives (velocity, acceleration, jerk and snap) over the trajectory.

\subsection{Endpoint cost function}
\label{sec:endpoint_cost}

The purpose of the endpoint cost function is to keep the local trajectory close to the global one. This is achieved by penalizing position and velocity deviation at the end of the optimized trajectory segment from the desired values that come from the global trajectory. Because the property is formulated as a soft constraint, the targeted values might not be achieved, for example, because of obstacles blocking the path. The function is defined as follows:

\begin{align}
E_{ep} &= \lambda_{p} (p(t_{ep}) - p_{ep})^2 + \lambda_{v} (p'(t_{ep}) - p'_{ep})^2,
\end{align}
where $t_{ep}$ is the end time of the segment, $p(t)$ is the trajectory to be optimized, $p_{ep}$ and $p'_{ep}$ are the desired position and velocity, respectively, $\lambda_{p}$ and $\lambda_{v}$ are the weighting parameters.

\subsection{Collision cost function}
\label{sec:collision_cost}

Collision cost penalizes the trajectory points that are within the threshold distance $\tau$ to the obstacles. The cost function is computed as the following line integral:
\begin{align}
E_{c} &= \lambda_{c} \int_{t_{min}}^{t_{max}} c(p(t)) ||p'(t)|| dt,
\end{align}
where the cost function for every point $c(x)$ is defined as follows:
\begin{align}
c(x) = 
\begin{cases}
\frac{1}{2\tau} (d(x)-\tau)^2 & \mbox{if }  d(x) \le \tau \\
0 & \mbox{if } d(x) > \tau,
\end{cases}
\end{align}
where $\tau$ is the distance threshold, $d(x)$ is the distance to the nearest obstacle, and $\lambda_{c}$ is a weighting parameter. The line integral is computed using the rectangle method, and distances to the obstacles are obtained from the precomputed EDT by using trilinear interpolation.

\subsection{Quadratic derivative cost function}
\label{sec:quadratic_cost}

Quadratic derivative cost is used to penalize the integral over square derivatives of the trajectory (acceleration, jerk, and snap), and is defined as follows:
\begin{align}
E_{ep} &= \sum_{i=2}^4 \int_{t_{min}}^{t_{max}} \lambda_{qi} (p^{(i)}(t))^2 dt.
\end{align}
The above function has a closed-form solution for trajectory segments represented using B-splines.

\subsection{Derivative limit cost function}
\label{sec:limit_cost}

\begin{figure}
\label{fig:limit_cost}
	    \centering
	    \includegraphics[width=0.5\linewidth]{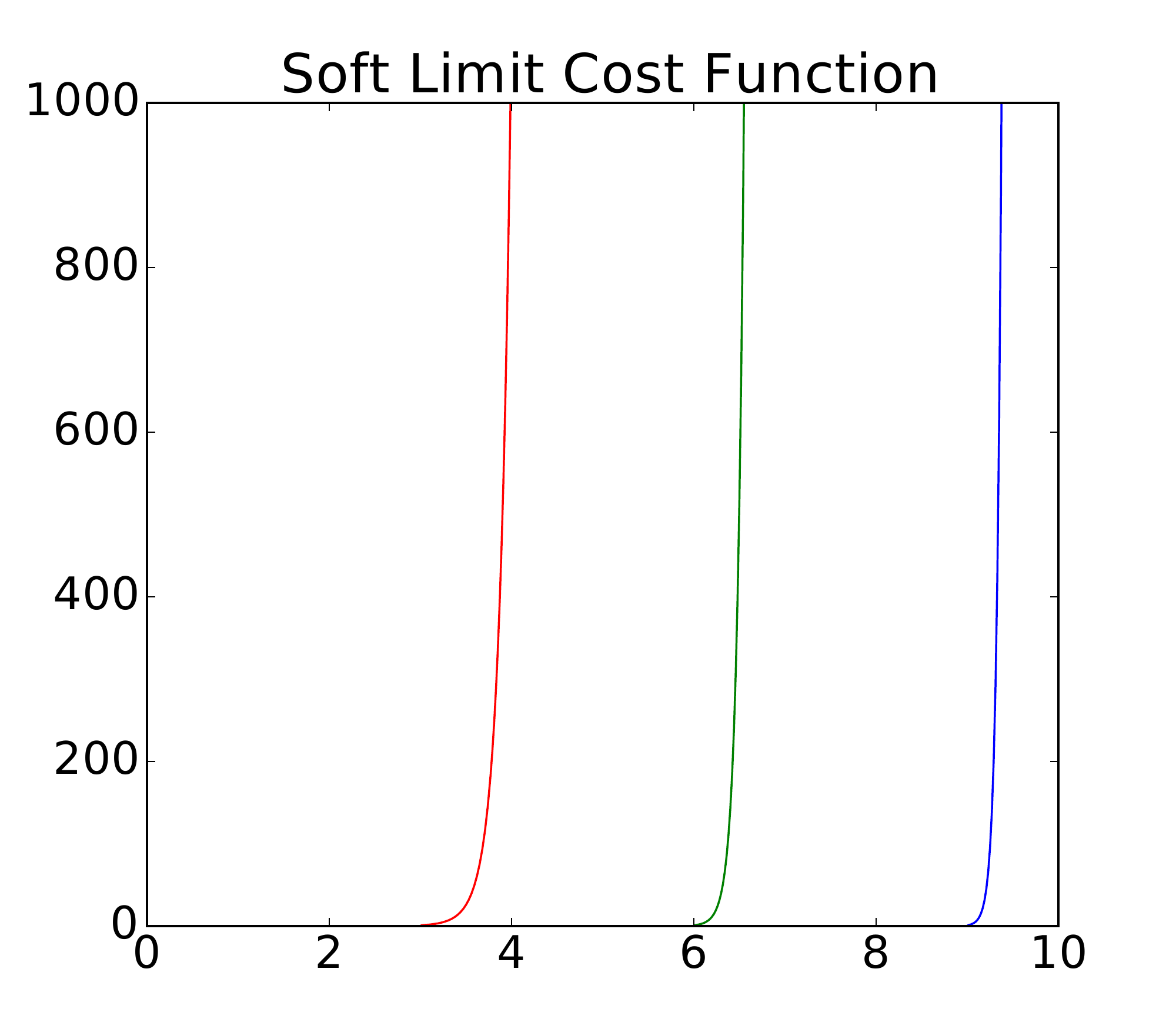}
        \caption{Soft limit cost function $l(x)$ proposed in Section \ref{sec:limit_cost} for $p_{max}$ equals three (red), six (green), and nine (blue). This function acts as a soft limit on the time derivatives of the trajectory (velocity, acceleration, jerk, and snap) to ensure they are bounded and are feasible to execute by the MAV. }
\end{figure}

To make sure that the computed trajectory is feasible, we must ensure that velocity, acceleration and higher derivatives of position remain bounded. This requirement can be included into the optimization as a constraint $\forall t : p^{(k)}(t) < p^k_{max}$, but in the proposed approach, we formulate it as a soft constraint by using the following function:
\begin{align}
E_{ep} &= \sum_{i=2}^4 \int_{t_{min}}^{t_{max}} l(p^{(i)}(t)) dt,
\end{align}
where $l(x)$ is defined as follows:
\begin{align}
l(x) = 
\begin{cases}
\exp((p^{(k)}(x))^2 - (p^{k}_{max})^2) -1 & \mbox{if } p^{(k)}(x) > p^{k}_{max} \\
0 & \mbox{if } p^{(k)}(x) \le p^{k}_{max}
\end{cases}
\end{align}

This allows us to minimize this cost function by using any algorithm designed for unconstrained optimization.

\subsection{Implementation details}
\label{sec:implementation_details}

To run the local replanning algorithm on the drone, we first compute a global trajectory by using the approach described in \cite{richter2016polynomial}. This gives us a polynomial spline trajectory that avoids all mapped obstacles.  Thereafter, we initialize our replanning algorithm with six fixed control points at the beginning of the global trajectory and $C$ control points that need to be optimized.

In every iteration of the algorithm we set the endpoint constraints (Sec. \ref{sec:endpoint_cost}) to be the position and velocity at $t_{ep}$ of the global trajectory. The collision cost (Sec. \ref{sec:collision_cost}) of the trajectory is evaluated using a circular buffer that contains measurements obtained using the RGB-D camera mounted on the drone. The weights of quadratic derivatives cost (Sec. \ref{sec:quadratic_cost}) are set to the same values as those used for global trajectory generation, and the limits (Sec. \ref{sec:limit_cost}) are set $20\%$ higher to ensure that the global trajectory is followed with the appropriate velocity while laterally deviating from it.

After optimization, the first control point from the points that were optimized is fixed and sent to the MAV position controller. Another control point is added to the end of the spline, which increases $t_{ep}$ and moves the endpoint further along the global trajectory.

For optimization we use \cite{nlopt}, which provides an interface to several optimization algorithms. We have tested the MMA \cite{svanberg2002class} and BFGS \cite{liu1989limited} algorithms for optimization, with both algorithms yielding similar performance.

\begin{table}
 \begin{tabular}{ m{3.4cm}  m{1.2cm} m{1.2cm}  m{1.2cm} } 
 \hline
 \centering \textbf{Algorithm} & \centering \textbf{Success Fraction} & \centering \textbf{Mean Norm. Path Length} & \textbf{Mean Compute time} [s] \\
 \hline
 \hline
Inf. RRT* + Poly & \textbf{0.9778} & \textbf{1.1946} & 2.2965 \\
RRT Connect + Poly & 0.9444 & 1.6043 & \textbf{0.5444} \\
\hline
CHOMP N = 10 & 0.3222 & 1.0162 & 0.0032 \\
CHOMP N = 100 & 0.5000 & 1.0312 & 0.0312 \\
CHOMP N = 500 & 0.3333 & 1.0721 & 0.5153 \\
\hline
\cite{oleynikova2016continuous-time} S = 2 jerk & 0.4889 & 1.1079 & \textbf{0.0310} \\
\cite{oleynikova2016continuous-time} S = 3 vel & 0.4778 & 1.1067 & 0.0793 \\
\cite{oleynikova2016continuous-time} S = 3 jerk & 0.5000 & 1.0996 & 0.0367 \\
\cite{oleynikova2016continuous-time} S = 3 jerk + Restart & \textbf{0.6333} & 1.1398 & 0.1724 \\
\cite{oleynikova2016continuous-time} S = 3 snap + Restart & 0.6222 & 1.1230 & 0.1573 \\
\cite{oleynikova2016continuous-time} S = 3 snap & 0.5000 & \textbf{1.0733} & 0.0379 \\
\cite{oleynikova2016continuous-time} S = 4 jerk & 0.5000 & 1.0917 & 0.0400 \\
\cite{oleynikova2016continuous-time} S = 5 jerk & 0.5000 & 1.0774 & 0.0745 \\
  \hline
Ours C = 2  &  0.4777 &  \textbf{1.0668} & \textbf{0.0008} \\
Ours C = 3  & 0.4777 &  1.0860 & 0.0011 \\
Ours C = 4  & 0.4888 &  1.1104 & 0.0015 \\
Ours C = 5  & 0.5111 &  1.1502 & 0.0021 \\
Ours C = 6  & 0.5555 &  1.1866 & 0.0028 \\
Ours C = 7  & 0.5222 &  1.2368 & 0.0038 \\
Ours C = 8  & 0.4777 &  1.2589 & 0.0054 \\
Ours C = 9  & \textbf{0.5777} &  1.3008 & 0.0072 \\
 \hline
\end{tabular}
\caption{Comparison of different path planning approaches. All results except thouse of the presented study are taken from \cite{oleynikova2016continuous-time}. Our approach performs similarly to approaches using polynomial splines without restarts, which indicates that B-splines can represent trajectories similar to those represented by polynomial splines. Lower computation times of our approach can be explained by the fact that unconstrained optimization occurs directly on the control points, unlike other approaches where the problem must first be transformed into an unconstrained form.}
\label{tab:comparison}
\end{table}

\section{Results}
\label{sec:results}

In this section, we present experimental results obtained using the proposed approach. First, we evaluate the mapping and the trajectory optimization components of the system separately for comparison with other approaches and justify their selection. Second, we evaluate the entire system in a realistic simulator in several different environments, and finally, present an evaluation of the system running on real hardware.

\subsection{Three-dimensional circular buffer performance}

We compare our implementation of the 3D circular buffer to the popular octree-based solution of \cite{hornung13auro}. Both approaches use the same resolution of 0.1 m. We insert the depth maps sub-sampled to the resolution of $160 \times 120$, which come from a real-world dataset \cite{sturm12iros}. The results (Fig. \ref{fig:octomap_ring_buffer}) show that insertion of the data is more than an order of magnitude faster with the circular buffer, but only a limited space can be mapped with this approach. Because we need the map of a bounded neighborhood around the drone for local replanning, this drawback is not significant for target application.

\subsection{Optimization performance}

To evaluate the trajectory optimization we use the forest dataset from \cite{oleynikova2016continuous-time}. Each spline configuration is tested in 9 environments with 10 random start and end positions that are at least 4 m away from each other. Each environment is $10 \times 10 \times 10$ $m^3$ in size and is populated with trees with increasing density. The optimization is initialized with a straight line and after optimization, we check for collisions. For all the approaches, the success fraction, mean normalized path length, and computation time are reported (Table \ref{tab:comparison}).

The results of the proposed approach are similar in terms of success fraction to those achieved with polynomial splines from \cite{oleynikova2016continuous-time} without restarts, but the computation times with the proposed approach are significantly shorter. This is because the unconstrained optimization employed herein directly optimizes the control points, while in \cite{oleynikova2016continuous-time}, a complicated procedure to transform a problem to the unconstrained optimization form \cite{richter2016polynomial} must be applied. 

Another example of the proposed approach for trajectory optimization is shown in Figure \ref{fig:optimization_example}, where a global trajectory is generated through a pre-defined set of points with an obstacle placed in the middle. The optimization is performed as described in Section \ref{sec:implementation_details}, with the collision threshold $\tau$ set to 0.5 m. As can be seen in the plot, the local trajectory in the collision free regions aligns with the global one, but when an obstacle is encountered, a smooth trajectory is generated to avoid it and ensure that the MAV returns to the global trajectory.

\begin{figure}
		\centering
                \includegraphics[width=0.7\linewidth]{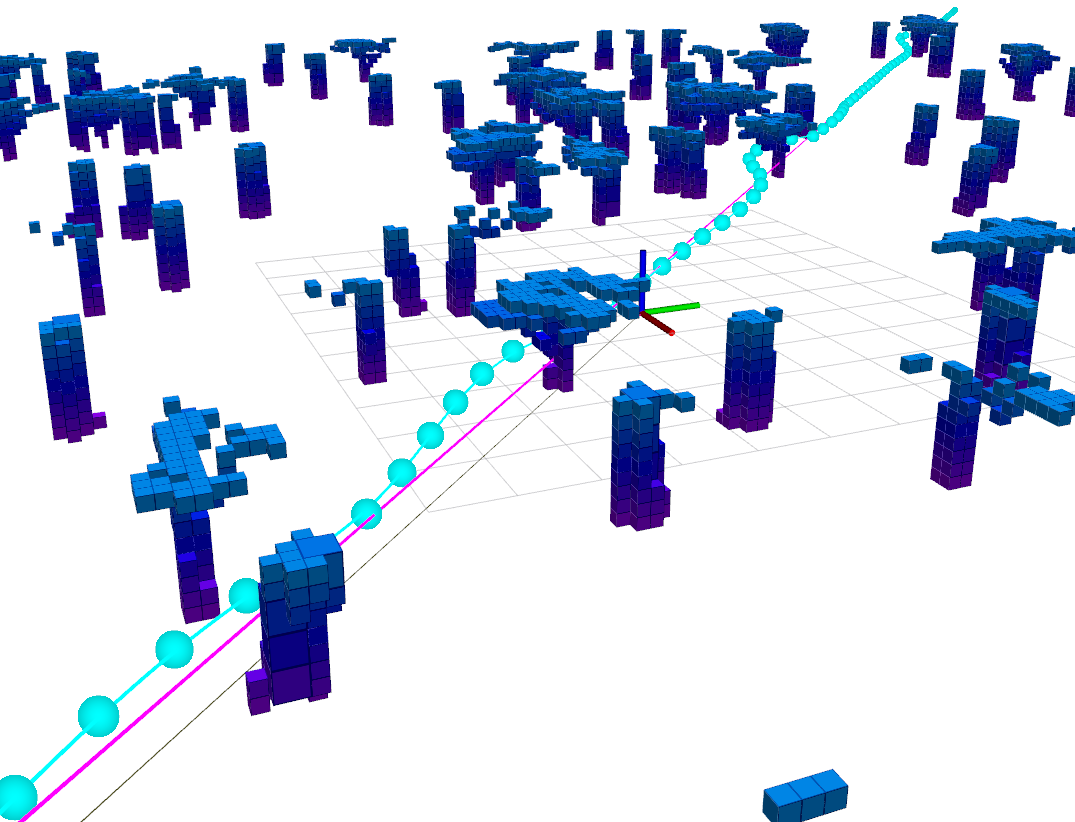}
        \caption{Result of local trajectory replanning algorithm running in a simulator on the forest dataset. The global trajectory is visualized in purple, local trajectory is represented as a uniform quintic B-spline, and its control points are visualized in cyan. Ground-truth octomap forest model is shown for visualization purposes.}
        \label{fig:simulation_big_forest}
\end{figure}

\subsection{System simulation}

To further evaluate our approach, we perform a realistic simulation experiment by using the Rotors simulator \cite{rotors:2016}. The main source of observations of the obstacles is a simulated RGB-D camera that produces  VGA depth maps at 20 FPS. To control the MAV, we use the controller developed by \citet{lee2010geometric}, which is provided with the simulator and modified to receive trajectory messages as control points for uniform B-splines. When there are no new commands with control points, the last available control point is duplicated and inserted into the B-spline. This is useful from the viewpoint of failure case because when an MAV does not receive new control points, it will slowly stop at the last received control point.

We present the qualitative results of the simulations shown in Figures \ref{fig:simulation} and \ref{fig:simulation_big_forest}. The drone is initialized in free space and a global path through the world populated with obstacles is computed. In this case, the global path is computed to intersect the obstacles intensionally. The environment around the drone is mapped by inserting RGB-D measurements into the circular buffer, which is then used in the optimization procedure described above.

In all presented simulation experiments, the drone can compute a local trajectory that avoids collisions and keeps it close to the global path. The timings of the various operations involved in trajectory replanning are presented in Table \ref{tab:timing}.

\subsection{Real-world experiments}

We evaluate our system on a multicopter in several outdoor experiments (Fig. \ref{fig:real_world_experiment}). In these experiments, the drone is initialized without prior knowledge of the map and the global path is set as a straight line with its endpoint in front of the drone 1 m above the ground. The drone is required to use onboard sensors to map the environment and follow the global path avoiding trees, which serve as obstacles.

We use the AscTec Neo platform equipped with a stereo camera for estimating drone motion and an RGB-D camera (Intel Realsense R200) for obstacle mapping. All computations are performed on the drone on a 2.1 GHz Intel i7 CPU.

In all presented experiments, the drone can successfully avoid the obstacles and reach the goal position. However, the robustness of the system is limited at the moment owing to the accuracy of available RGB-D cameras that can capture outdoor scenes.

\begin{table}
 \begin{tabular}{ m{1.2cm} | m{1.0cm} m{1.0cm} m{1.0cm} m{1.0cm} m{1.0cm} } 
 \hline
 \centering \textbf{Operation} & \centering Computing 3D points & \centering Moving volume & Inserting measurements & SDF computation & Trajectory optimization \\
 \hline
 \textbf{Time [ms]} & 
 0.265 & 0.025 & 0.518 &  9.913 & 3.424 \\
 \hline
\end{tabular}
\caption{Mean computation time for operations involved in trajectory replanning in the simulation experiment with depth map measurements sub-sampled to $160 \times 120$ and seven control points optimized.}
\label{tab:timing}
\end{table}

\section{Conclusion} 
\label{sec:conclusion}

In this paper, we presented an approach to real-time local trajectory replanning for MAVs. We assumed that the global trajectory computed by an offline algorithm is provided and formulated an optimization problem that replans the local trajectory to follow the global one while avoiding unmodeled obstacles.

We improved the optimization performance by representing the local trajectory with uniform B-splines, which allowed us to perform unconstrained optimization and reduce the number of optimized parameters.

For collision checking we used the well-known concept of circular buffer to map a fixed area around the MAV, which improved the insertion times by an order of magnitude compared to those achieved with an octree-based solution.

In addition, we presented an evaluation of the complete system and specific sub-systems in realistic simulations and on real hardware.

\section*{Acknowledgments}

This work has been partially supported by grant CR 250/9-2 (Mapping
on Demand) of German Research Foundation (DFG) and grant 608849 (EuRoC) of European Commission FP7 Program.

We also thank the authors of \cite{oleynikova2016continuous-time} for providing their dataset for evaluation of the presented method.


\bibliographystyle{plainnat}
\bibliography{references}

\end{document}